\documentclass[letterpaper, 10 pt, conference]{ieeeconf}  % Comment this line out if you need a4paper

\IEEEoverridecommandlockouts                              % This command is only needed if 
\usepackage{epsfig}         % for postscript graphics files
\usepackage{amsmath}        % assumes amsmath package installed
\usepackage{amssymb}        % assumes amsmath package installed
\usepackage{graphicx}
\usepackage{pgfplots}
\pgfplotsset{compat=1.18} 
\usepackage{cite}
\usepackage{tabularx} % 设定表格宽度
\usepackage{booktabs}
\usepackage{url}

\title{\LARGE \bf
HL-IK: A Lightweight Implementation of Human-Like Inverse Kinematics in Humanoid Arms
}

\author{Bingjie Chen$^{\dagger}$, Zihan Wang$^{\dagger}$, Zhe Han$^{\dagger}$, Guoping Pan, Yi Cheng, Chenxi Han, Houde Liu$^*$}

\begin{document}
\maketitle
\thispagestyle{empty}
\pagestyle{empty}

%%%%%%%%%%%%%%%%%%%%%%%%%%%%%%%%%%%%%%%%%%%%%%%%%%%%%%%%%%%%%%%%%%%%%%%%%%%%%%%%%%%%%%%%%%%%%%%%%%%%%%%%%
\begin{abstract}
Traditional IK methods for redundant humanoid manipulators emphasize end-effector (EE) tracking, frequently producing configurations that are valid mechanically but not human-like. We present Human-Like Inverse Kinematics (HL-IK), a lightweight IK framework that preserves EE tracking while shaping whole-arm configurations to appear human-like—without full-body sensing at runtime. The key idea is a learned elbow prior: using large-scale human motion data retargeted to the robot, we train a FiLM-modulated spatio-temporal attention network (FiSTA) to predict the next-step elbow pose from the EE target and a short history of EE–elbow states. This prediction is incorporated as a small residual alongside EE and smoothness terms in a standard Levenberg–Marquardt optimizer, making HL-IK a drop-in for numerical IK frameworks given a defined elbow frame and pose interface. Over 183k simulation steps, HL-IK reduces arm-similarity position and direction error by 30.6\% and 35.4\% on average, and by 42.2\% and 47.4\% on the most challenging trajectories. Hardware teleoperation on a robot distinct from simulation further confirms the gains in anthropomorphism. HL-IK is simple to integrate, adaptable across platforms via our pipeline, and adds minimal computation, enabling human-like motions for humanoid robots. %Project page: \url{https://hl-ik.github.io/}

\end{abstract}

%%%%%%%%%%%%%%%%%%%%%%%%%%%%%%%%%%%%%%%%%%%%%%%%%%%%%%%%%%%%%%%%%%%%%%%%%%%%%%%%%%%%%%%%%%%%%%%%%%%%%%%%%
\section{Introduction}
A robotic arm can be defined as a series of links connected together by joints \cite{Paul}. Inverse kinematics (IK) is a fundamental problem in such robotics, traditionally formulated to compute joint configurations that achieve a specified end-effector (EE) pose. For industrial manipulators, this formulation is often sufficient, since the primary objective is to place the tool center point at the desired location with high precision. Classical IK solvers—whether based on closed-form derivations \cite{Tian, Zheng, Vu, Diankov}, numerical iterations \cite{Murray, Yonezawa}, or optimization frameworks \cite{Santos, Maric, Sundaralingam} — focus almost exclusively on EE tracking.

For redundant robotic arms, the inverse solution to a given EE pose is often not unique, with infinitely many possible configurations \cite{Ames, Dione}. When only the EE pose is constrained, the intermediate joints remain underdetermined \cite{Boukheddimi}, which can lead to solutions that are mechanically valid but visually unnatural and non-human-like. In scenarios such as humanoid robot teleoperation \cite{Darvisha, Seoa, Handa}, beyond accurate EE tracking, we also aim for the robot’s overall arm configuration to closely resemble that of the human arm, thereby achieving a higher level of anthropomorphism. Existing methods \cite{Heb, Zhou} often rely on external cameras to capture human body keypoints and align them with robot joints to improve configuration similarity. Yet, such approaches not only require additional perception inputs but also typically do not treat EE tracking as the primary constraint, and thus cannot be regarded as strict IK solutions. Therefore, our goal is to develop a system that, given only the desired EE pose as input (as in traditional IK), not only ensures precise EE tracking but also achieves close similarity between the human and robot arm configurations. 

% Common approaches are to use rules such as obstacle avoidance constraints \cite{Rakita, Tenhumberg, Kang} or minimizing joint movement to select an optimal solution. However, in certain scenarios—such as the teleoperation of humanoid robots \cite{Darvisha, Seoa} — it is not enough to ensure accurate EE tracking. We also aim for the robot’s overall arm configuration to closely resemble that of the human arm, thereby achieving a higher level of anthropomorphism.

\begin{figure}
    \centering
    \includegraphics[width=0.48\textwidth]{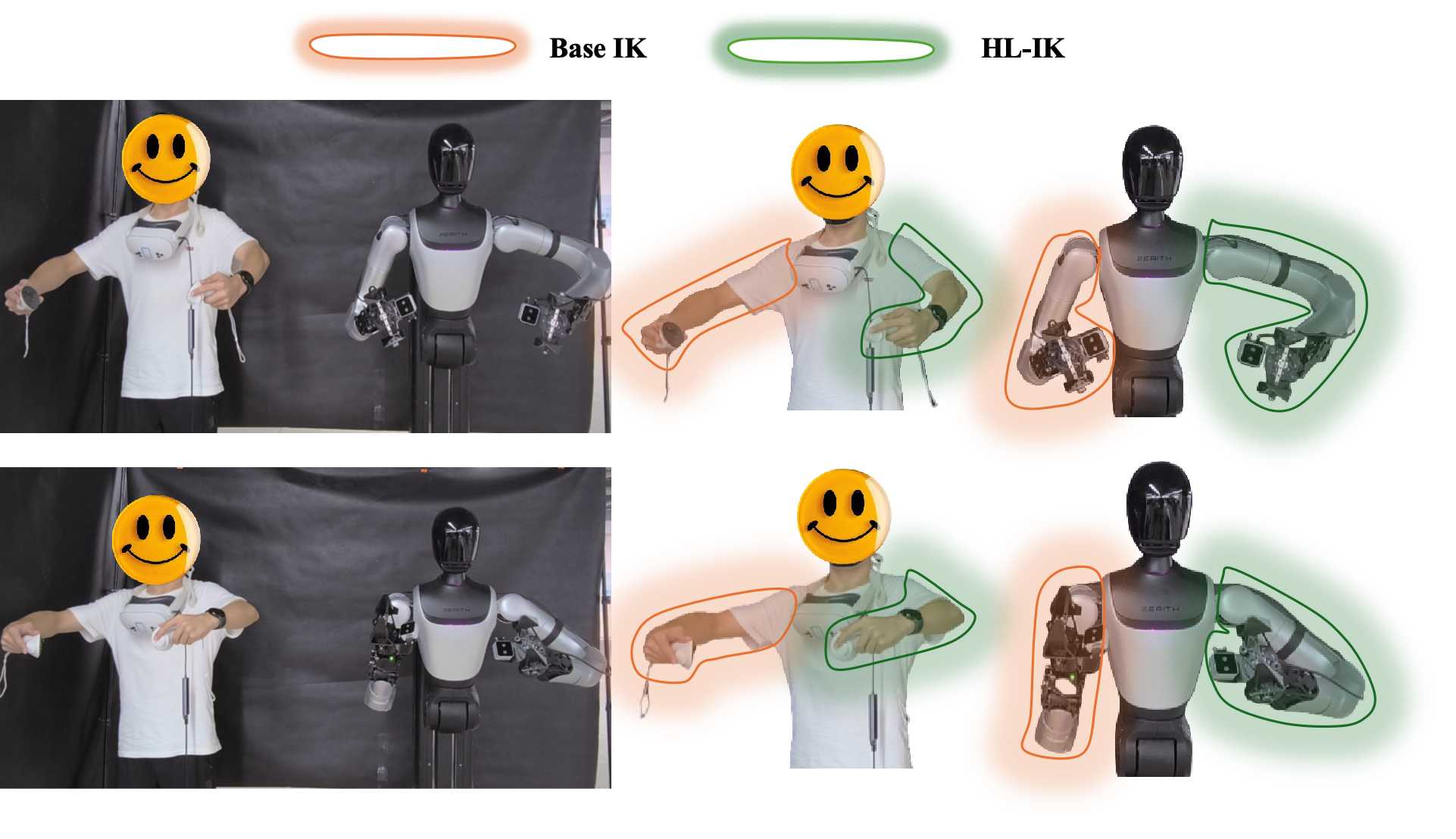}
    \caption{In physical teleoperation, when only the EE pose is used as input, our HL-IK method achieves a more human-like effect, with the arm configuration more closely resembling that of humans.}
    \label{fig:show}
\end{figure}

To realize this goal, we first model the human arm as a four-point, three-segment kinematic chain comprising the shoulder, elbow, wrist, and fingertips (the human EE) [19]. For a fixed EE pose, the dominant redundancy manifests as the elbow ``swivel" about the shoulder–wrist axis. Aligning the elbow pose effectively sets the arm plane and the forearm pointing direction, thereby resolving the main ambiguity and yielding anthropomorphic configurations without sacrificing EE accuracy. In other words, once the elbow is aligned, the overall arm configuration becomes perceptually natural and significantly more similar to that of a human. Furthermore, given a desired EE pose, determining a prior elbow pose that best reflects the natural human form becomes a central aspect of our approach. In summary, the primary contributions of this paper are:

1) Human-like data acquisition framework: We propose an automatic EE–elbow data collection scheme based on large-scale human motion trajectory datasets, which can be readily adapted to different robots.

2) Elbow prediction network: We design a FiLM-modulated Spatio-Temporal Attention Network (FiSTA) that uses only a partial history of EE and elbow frames to predict the desired human-like elbow pose for a given EE target.

3) Human-like IK (HL-IK) framework: We build an end-to-end HL-IK framework that (i) takes only EE targets as input, (ii) predicts a human-like elbow pose to resolve redundancy, and (iii) injects it as a residual cost into numerical IK, while keeping end-effector tracking as the primary objective and maintaining low runtime cost. Fig.~\ref{fig:show} clearly illustrates the improved anthropomorphism achieved by our method.

\begin{figure*}[t]
	\centering
	\includegraphics[width=0.84\textwidth]{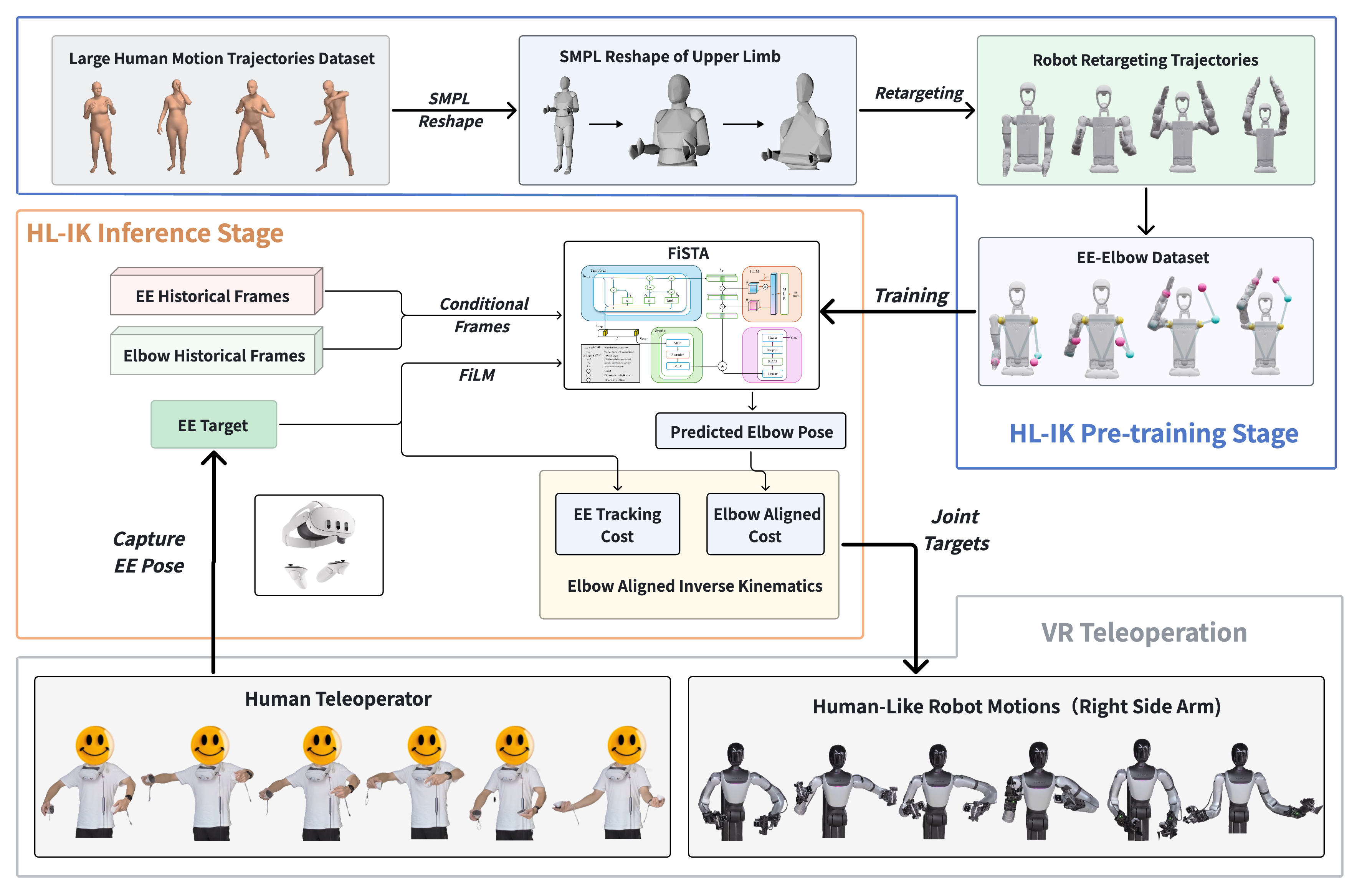}
	\caption{Our method begins with a large-scale human motion dataset, which we retarget to the robot to extract an EE–elbow mapping dataset used to train our FiSTA network. After training, during human teleoperation we obtain the operator’s desired current EE pose via a VR. Given this target and a fixed-length history of past frames, FiSTA predicts the elbow position. We then augment the IK objective with an elbow-alignment cost and solve for the desired joint angles using Levenberg–Marquardt iterations. The resulting commands are sent to the robot’s low-level controller.}
\label{pipeline}
\end{figure*}

\section{Related Work}
\subsection{Inverse Kinematics}
The goal of IK is to compute joint configurations that realize a specified EE pose. Classical approaches pursue analytical (closed-form) solutions \cite{Tian, Zheng, Vu}, which are highly efficient but hinge on restrictive assumptions about the manipulator’s kinematic structure. More recently, the field has shifted toward iterative optimization formulations \cite{Santos, Maric, Sundaralingam, Wangb} that trade closed-form speed for modeling flexibility and can accommodate secondary objectives arising from kinematic redundancy, including geometric constraints \cite{Sundaralingam, Rakitaa} and learned collision penalties \cite{Rakita}. Building on this line of work, our proposed HL-IK explicitly incorporates humanoid-likeness into the IK objective. While ensuring precise EE tracking, HL-IK further guides the arm configuration to resemble natural human arm postures, thereby achieving more anthropomorphic motion behavior.

% \subsection{Human-Like Motion Planning}
% Human-like motion planning aims to generate robot motions that both achieve the desired end-effector behavior and resemble natural human arm configurations, and it is closely related to motion retargeting—mapping motions from a source entity to a target entity, such as transferring human movements to robots \cite{Gleicher}. This capability is crucial for robotic teleoperation systems \cite{Schmidt} and for learning frameworks based on human demonstrations \cite{Heb, Fua, Zhangc}. Existing real-time retargeting-based teleoperation often requires full-body sensing and additional perception to recover the complete human state, whereas HL-IK instead pretrains on large-scale retargeting data to learn an elbow-configuration prior, enabling human-like upper-limb motion from end-effector-only input. Unlike priors that are tightly coupled to closed-form solvers, we learn an optimizer-agnostic elbow prior and encode it as a cost term. Because it is decoupled from any specific solver, the prior can be added as a generic objective component and plugged directly into standard numerical optimization pipelines, offering substantially greater flexibility.

\subsection{Motion Retargeting}
Motion retargeting refers to mapping motions from a source entity to a target entity \cite{Gleicher}, such as transferring human movements to robots. It plays a crucial role in robotic teleoperation systems \cite{Schmidt} and in learning frameworks based on human demonstrations \cite{Heb, Fua, Zhangc}. Existing real-time whole-body teleoperation methods based on motion retargeting rely on sensing devices to capture the full-body state, which in turn requires additional perception algorithms. In contrast, HL-IK pretrains on large-scale retargeting data to learn an elbow-configuration prior, enabling human-like upper-limb motion from EE-only input.

\subsection{Human-Like Motion Planning in IK}
To encourage human-like arm configurations in IK, prior work largely takes two paths. Classical methods impose kinematic heuristics—e.g., swivel-angle parameterizations \cite{Kimc, Liuc} and physiology-inspired criteria \cite{Zhenga}—but such hand-tuned rules generalize poorly across subjects and tasks. We instead learn an optimizer-agnostic elbow prior from large-scale human trajectories and encode it as a lightweight cost. Decoupled from any specific solver, this prior drops in as a generic objective term, integrating seamlessly with standard numerical optimizers.

\section{Methodology}
To realize HL-IK, we first retarget large-scale human motion datasets onto the robot to collect and process paired EE-elbow trajectories. We then introduce our FiLM-modulated Spatio-Temporal Attention (FiSTA) network, which learns from these data to predict the elbow pose given a short motion history and the current desired EE target. Finally, we show how the predicted elbow is incorporated into a numerical optimization IK. The full pipeline of our approach is illustrated in Fig. \ref{pipeline}.

\subsection{EE-Elbow Data Collection}

\begin{figure}
	\centering
	\includegraphics[width=0.48\textwidth]{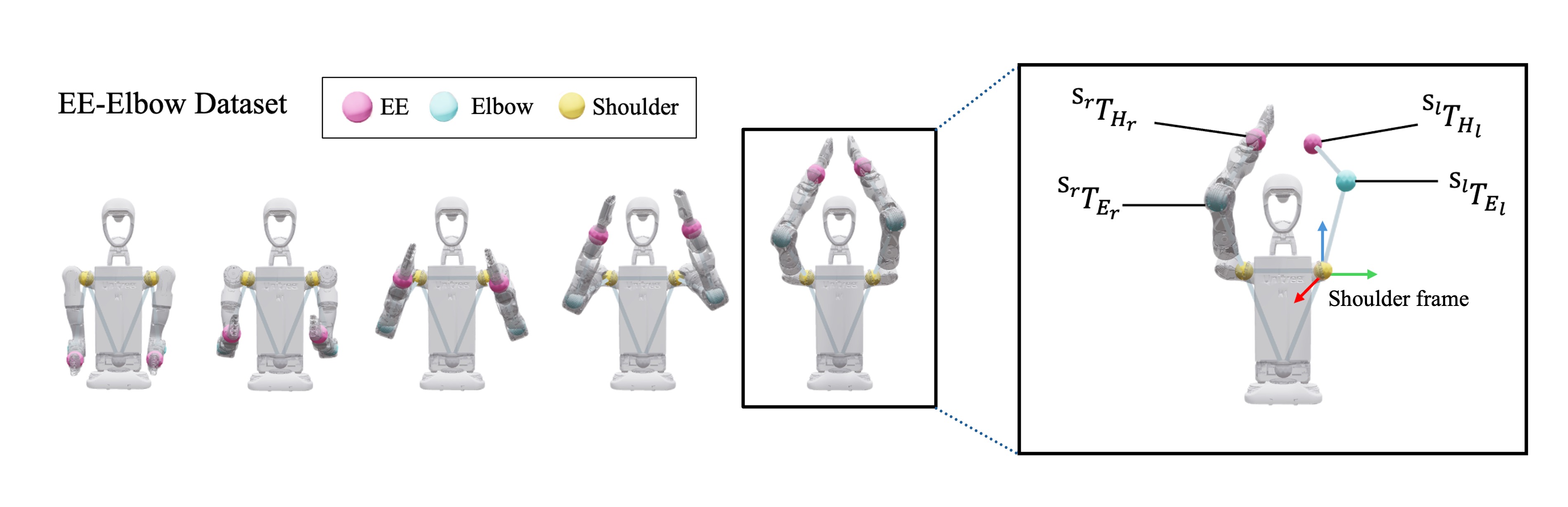}
	\caption{EE-Elbow Data Collection. For each frame, extract the relative pose of the EE and the elbow in the corresponding shoulder frames.}
\label{data_collection}
\end{figure}

We use AMASS \cite{AMASS:ICCV:2019} as the source of human motion trajectories, which provides consistent Skinned Multi-Person Linear model (SMPL)  sequences with smooth kinematics and reliable pose quality. We retain six upper-limb joints—left/right shoulder, elbow, and EE—to represent the two arms. Following \cite{Heb}, we first estimate an upper-body shape parameter \(\beta'\) for the target robot via gradient-based fitting. Using \(\beta'\), we retarget each AMASS frame to the robot and build a frame-aligned dataset \(\mathcal{D}_{\beta'}\) paired with the original AMASS sequence \(\mathcal{D}_{\beta}\).

We adopt the convention that \({}^{A}\mathbf{T}_{B}\in SE(3)\) maps coordinates from frame \(\{B\}\) to frame \(\{A\}\). All poses in \(\mathcal{D}_{\beta'}\) are expressed in the world frame \(\{W\}\). From the retargeted sequence, forward kinematics (using the robot URDF) yields the world transforms of the shoulder, elbow, and hand (EE) for each arm:
\({}^{W}\mathbf{T}_{S_j}\), \({}^{W}\mathbf{T}_{E_j}\), \({}^{W}\mathbf{T}_{H_j}\) with \(j\in\{\mathrm{left},\mathrm{right}\}\).
The quantities we extract are the elbow and EE poses in the corresponding shoulder frames:
\begin{subequations}
\begin{gather}
{}^{S_j}\mathbf{T}_{E_j} = \big({}^{W}\mathbf{T}_{S_j}\big)^{-1}\,{}^{W}\mathbf{T}_{E_j}, \\
{}^{S_j}\mathbf{T}_{H_j} = \big({}^{W}\mathbf{T}_{S_j}\big)^{-1}\,{}^{W}\mathbf{T}_{H_j}.
\end{gather}
\end{subequations}

These four \(SE(3)\) poses per frame—\({}^{S_{\mathrm{l}}}\mathbf{T}_{E_{\mathrm{l}}}\), \({}^{S_{\mathrm{l}}}\mathbf{T}_{H_{\mathrm{l}}}\), \({}^{S_{\mathrm{r}}}\mathbf{T}_{E_{\mathrm{r}}}\), \({}^{S_{\mathrm{r}}}\mathbf{T}_{H_{\mathrm{r}}}\)—constitute the supervision we use for learning and downstream IK. Fig.~\ref{data_collection} illustrates this data collection process.

\subsection{FiLM-modulated Spatio-Temporal Attention Network for Elbow Prediction}
\begin{figure}
    \centering
\includegraphics[width=0.48\textwidth]{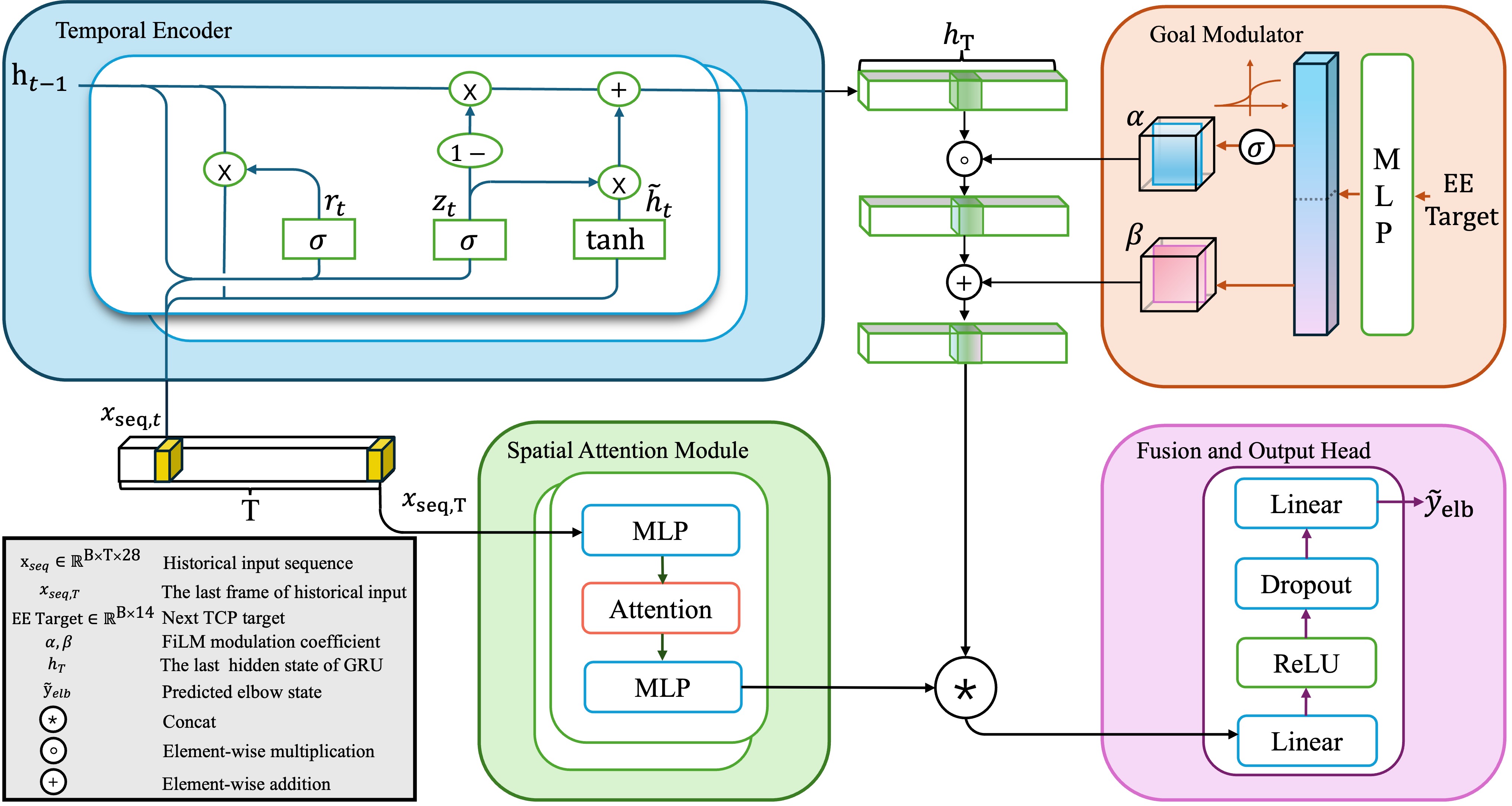}
    \caption{Model Architecture. A GRU encodes the 5-frame history to produce temporal features, which are FiLM-conditioned using the next EE target. In parallel, an attention module computes spatial features from the last frame. The two streams are then concatenated to predict the elbow configuration.}
    \label{fig:model_architecture}
\end{figure}

\textbf{Inputs and outputs.} For a single arm, the history window contains \(T\) frames. Each frame encodes the \(7\)D pose (3D position \(+\) 4D unit quaternion) of the EE and the elbow in the shoulder-local frame, giving \(7 \times 2 \times T = 14T\) dimensions in total.
The future conditioning input is the next-step EE target pose (7D).
The network predicts a 7D elbow pose for the next step.

\textbf{FiSTA architecture.} Fig.~\ref{fig:model_architecture} illustrates the network architecture of FiSTA and the model comprises four cooperating modules:

\subsubsection{Temporal Encoder}
A GRU consumes the 5-frame history and produces a temporal summary that captures position, velocity changes, and inertial trends.

\subsubsection{Spatial Attention Module}
On the most recent frame, the EE pose and elbow pose are treated as two tokens; a lightweight self-attention layer models their instantaneous dependence to yield a spatial summary.

\subsubsection{Goal Modulator}
The next-step EE target is mapped to Feature-wise Linear Modulation (FiLM)-style \cite{perez2018film} feature-wise scale and shift, which affine-modulate the temporal summary and thus condition historical dynamics on the imminent goal.

\subsubsection{Fusion and Output Head}
The modulated temporal features are concatenated with the spatial summary and passed through an Multi-Layer Perceptron(MLP) regressor to produce the 7D elbow prediction.

\textbf{Design rationale.}
This design explicitly decouples temporal dynamics from instantaneous spatial coordination while enabling goal-conditioned modulation.
Compared with naive target concatenation, FiLM modulation provides more expressive conditioning, and the lightweight spatial attention captures crucial EE--elbow coordination with negligible computational overhead.

\subsection{Elbow Aligned Inverse Kinematics}
Kinematic optimization benefits from quasi-Newton approaches, which accelerate convergence by exploiting curvature information of the cost function \cite{PyRoki}. In our IK computation, we employ the Levenberg–Marquardt (LM) optimizer \cite{Levenberg, Marquardt, Wampler, Nakamura, Buss}, which minimizes the sum of squared residuals defined below.

\subsubsection{EE Pose Cost} This EE pose cost $\boldsymbol{c}_{\text{ee}}$ allows the IK to track the given EE target as closely as possible.
\begin{equation}
    \boldsymbol{c}_{\text{ee}}(\boldsymbol{q})
    = \mathbf{W}_{\text{ee}}^{1/2}\,
    \log \left( \left(\mathbf{T}_{\text{base}}^{\text{tar}_\text{ee}}\right)^{-1}
    \mathbf{T}_{\text{base}}^{\text{ee}}(\boldsymbol{q}) \right) \in \mathbb R^{6},
\end{equation}

\subsubsection{Elbow Pose Cost} This elbow pose cost $\boldsymbol{c}_{\text{elbow}}$ enables the IK to track the given predicted elbow pose to a certain extent.\textbf{}
\begin{equation}
    \boldsymbol{c}_{\text{elbow}}(\boldsymbol{q})
    = \mathbf{W}_{\text{elbow}}^{1/2}\,
    \log \left( \left(\mathbf{T}_{\text{base}}^{\text{tar}_\text{el}}\right)^{-1}
    \mathbf{T}_{\text{base}}^{\text{el}}(\boldsymbol{q}) \right) \in \mathbb R^{6}
\end{equation}

\subsubsection{Smoothness Cost} The smoothness cost $\boldsymbol{c}_{smooth}$ encourages small changes in joint positions, and is useful for generating smooth trajectories.
\begin{equation}
    \boldsymbol{c}_{\text{smooth}}(\boldsymbol{q})
    = \mathbf{W}_{\text{smooth}}^{1/2}\,
    \left(\boldsymbol{q}_{t}-\boldsymbol{q}_{t-1}\right) \in \mathbb R^{d}
\end{equation}

\noindent
where 
\begin{itemize}
    \item $\mathbf{T}_{\text{base}}^{\text{tar}_\text{ee}} \in SE(3)$ is the target EE pose expressed in the base frame,
    \item $\mathbf{T}_{\text{base}}^{\text{ee}}(\boldsymbol{q}) \in SE(3)$ is the forward-kinematics EE pose at configuration $\boldsymbol{q}$,
    \item $\mathbf{T}_{\text{base}}^{\text{tar}_\text{el}} \in SE(3)$ is the target elbow pose in the base frame,
    \item $\mathbf{T}_{\text{base}}^{\text{el}}(\boldsymbol{q}) \in SE(3)$ is the elbow pose from forward kinematics,
    \item $\log(\cdot): SE(3) \rightarrow \mathbb{R}^{6}$ is the matrix logarithm mapping rigid-body transformations to 6D twist coordinates (translation + rotation error),
    \item $\mathbf{W}_{(\cdot)}^{1/2}$ denotes the square-root of the weight matrix used for residual scaling,
    \item $\boldsymbol{q}_t \in \mathbb{R}^{d}$ is the robot joint configuration at time step $t$ and the total number of joints is $d$.
\end{itemize}

Then, the three types of residuals are stacked into a single residual vector $\boldsymbol{c}(\boldsymbol{q})$, and its Jacobian $\mathbf{J}(\boldsymbol{q})$ with respect to the joint configuration $\boldsymbol{q}$ is computed. The Levenberg–Marquardt iteration then updates the joint configuration by solving a damped least-squares problem, which balances fast convergence with robustness against ill-conditioning:
\begin{subequations}
\begin{gather}
\boldsymbol{c}(\boldsymbol{q}) =
\begin{bmatrix}
\boldsymbol{c}_{\text{ee}}(\boldsymbol{q}) \\
\boldsymbol{c}_{\text{elbow}}(\boldsymbol{q}) \\
\boldsymbol{c}_{\text{smooth}}(\boldsymbol{q})
\end{bmatrix}, \\[4pt]
\mathbf{J}(\boldsymbol{q})=\frac{\partial \boldsymbol{c}(\boldsymbol{q})}{\partial \boldsymbol{q}}, \\[4pt]
\boldsymbol{q}_{n}
= \boldsymbol{q}_{n-1} - \left(\mathbf{J}^{\top}\mathbf{J} + \lambda \mathbf{I} \right)^{-1}
\mathbf{J}^{\top} \boldsymbol{c}(\boldsymbol{q}_{n-1}).
\end{gather}
\end{subequations}

Through successive iterations, the joint configuration $\boldsymbol{q}_n$ converges to a solution $\boldsymbol{q}^\ast$ that minimizes the stacked costs, yielding an IK solution with accurate EE tracking, improved elbow alignment, and smooth joint motions.

\section{Experiments}

\subsection{Model Evaluation}

To comprehensively assess the elbow–prediction network, we conduct systematic experiments along three axes: (i) Network Comparisons; (ii) Tests on history length and ablation experiments on the FiSTA network; (iii) Runtime tests. To ensure broad coverage—from everyday activities to dynamic, in the AMASS dataset, we specifically employ the following three items:
\begin{enumerate}[]
    \item ACCAD Dataset, 252 trajectories \cite{AMASS_ACCAD}
    \item CMU Dataset, 2079 trajectories \cite{AMASS_CMU}
    \item SFU Dataset, 44 trajectories \cite{AMASS_SFU}
\end{enumerate}
To balance efficiency and reliability, we adopt a two-stage training strategy: (1) All network comparison experiments, ablation studies, and hyperparameter searches are conducted on the ACCAD Dataset. (2) We then perform final training on the full dataset using the selected configuration. During comparison, we split the dataset by trajectory, with 90\% for training and 10\% for validation. All experiments are repeated with four random seeds (42, 123, 789, 1024) to ensure statistical reliability, and we report mean $\pm$ standard deviation.

\begin{figure}[t]
    \centering
    \includegraphics[width=0.45\textwidth]{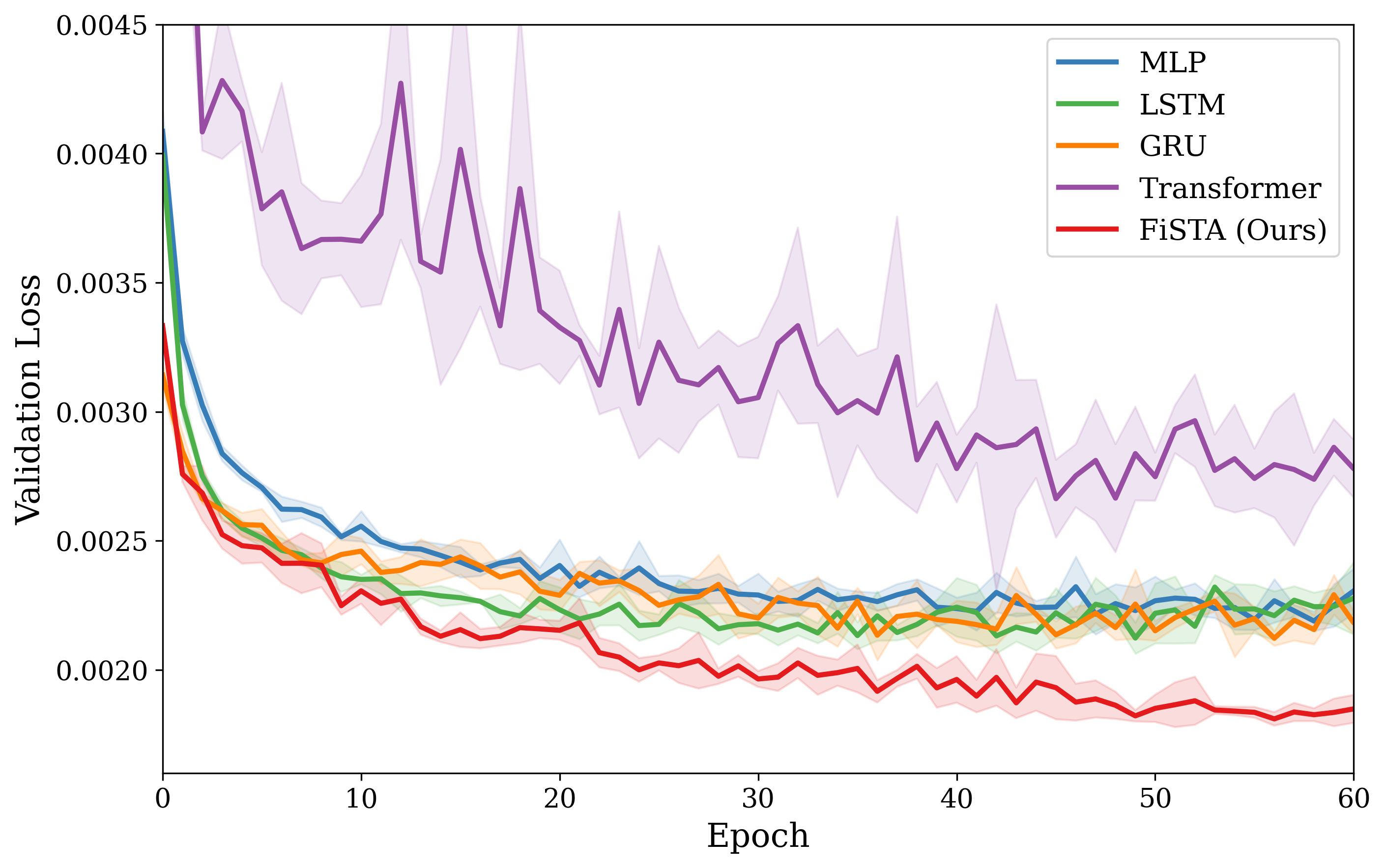}
    \caption{Model Evaluation Comparison. The y-axis shows validation MSE over epochs. The red star marks the lowest loss of FiSTA. All models use a 5-frame history, with other hyperparameters set to their best configuration.}
    \label{fig:arch_compare}
\end{figure}

\subsubsection{Network Comparisons}
We evaluated five representative architectures—MLP, Long Short-Term Memory (LSTM), Gated Recurrent Unit (GRU), Transformer, and FiSTA—using the ACCAD subset, where each was trained with tuned hyperparameters under identical normalization and loss settings. The network is trained by minimizing the Mean Squared Error (MSE) and all baseline models share the same input-output specification as FiSTA. The history length was fixed at 5 for all models. Validation errors are reported in Fig.~\ref{fig:arch_compare}. Among the candidates, FiSTA achieved the lowest validation error (\texttt{MSE=0.001781 $\pm$ 0.000014}), substantially outperforming all baselines. By contrast, the explicit architectural separation of spatial and temporal processing in FiSTA provides a stronger inductive bias than standard recurrent models or Transformers, which must learn these relationships from a less structured hidden state.

\begin{figure}
    \centering
    \includegraphics[width=0.45\textwidth]{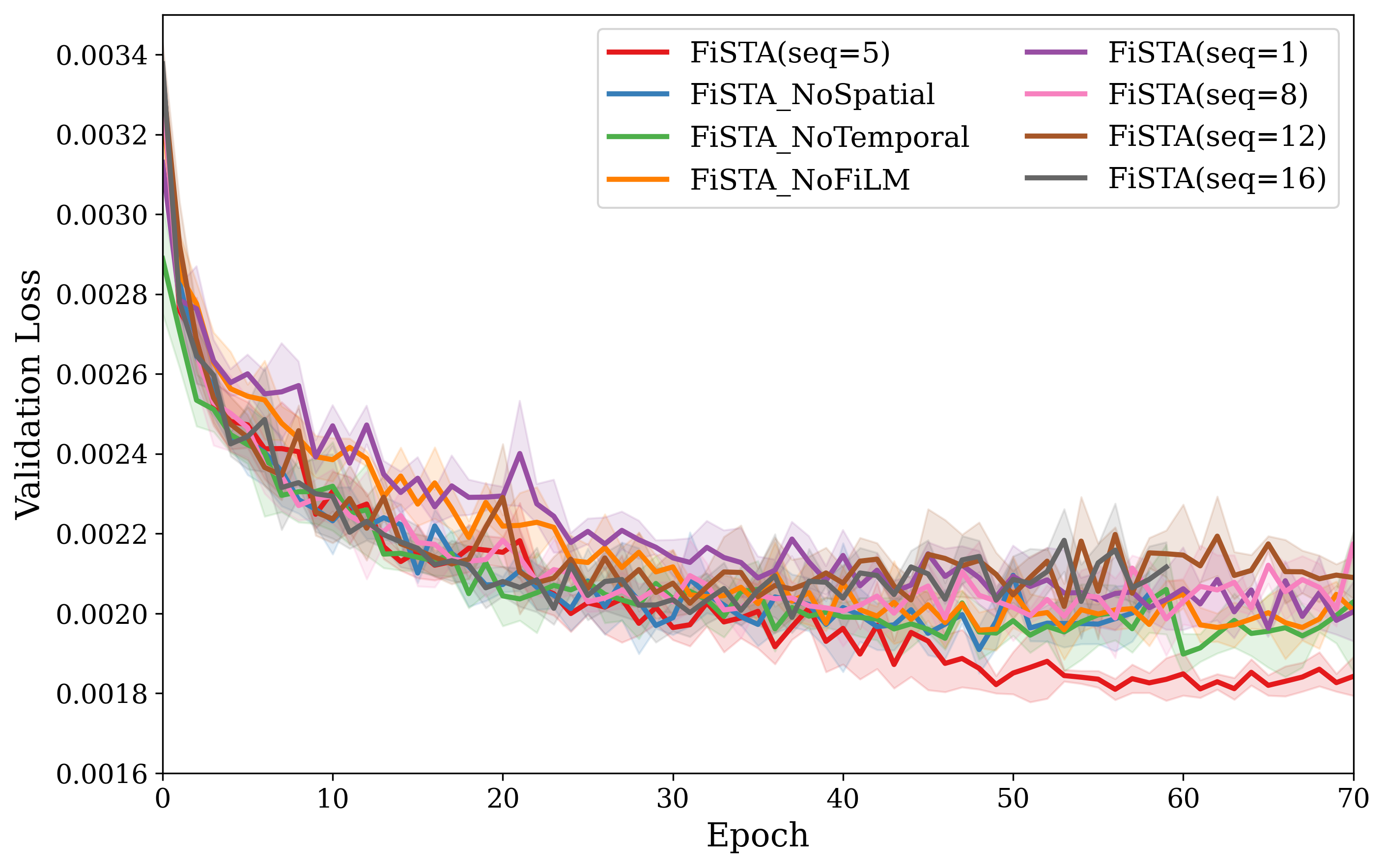}
    \caption{Ablation and sequence length comparison of FiSTA. Validation loss with different history lengths (1, 5, 8, 12, 16 frames) and ablation variants (without spatial, FiLM, or temporal). Shaded regions indicate standard deviation across four random seeds.}
    \label{fig:hist_len}
\end{figure}

\subsubsection{Tests on history length and ablation experiments on the FiSTA network}
To rigorously evaluate our proposed model, we conducted a series of tests to analyze the impact of history sequence length and core architectural components on the final validation loss. First, we assessed the model's performance with varying lengths of historical context, testing sequence lengths of 1, 5, 8, 12, and 16 frames. As shown in Fig.~\ref{fig:hist_len}, the results clearly indicate an optimal contextual window. The model with a sequence length of 5 achieved the lowest validation loss, demonstrating maximum predictive precision. We observed that $L=5$ strikes a favorable balance: too short a context fails to provide sufficient dynamic information for inferring motion trends, while an excessively long context introduces irrelevant historical data, reducing the signal-to-noise ratio. To quantify the contribution of each component, we performed an ablation study on \texttt{FiSTA (seq=5)} (all deltas relative to the full model). Removing the Spatial Attention and Temporal modules increased the loss by 4.5\% and 2.3\%, respectively, indicating that both are important for building a robust spatiotemporal representation. The largest effect came from ablating the Goal Modulation module, which led to a 6.0\% drop in accuracy. This module, implemented as a FiLM layer, conditions the GRU-encoded temporal features on the EE target. Its removal therefore highlights the critical role of FiLM-based goal conditioning in injecting target information into the temporal feature stream for this task.

\begin{table*}[t]
\centering
\caption{The Average runtime and the minimum loss comparison of different models}
\label{tab:final_runtime_comparison}
\setlength{\tabcolsep}{8pt}
\begin{tabular}{l c c c c} 
\hline
\textbf{Model Name} & \textbf{Preprocess (ms)} & \textbf{IK (ms)} & \textbf{Total (ms)} & \textbf{Loss ($\times 10^{-3}$)} \\
\hline\hline
MLP                  & \textbf{1.3104 $\pm$ 0.153} & 5.4896 $\pm$ 0.951 & 6.8000 $\pm$ 0.963 & 2.106 $\pm$ 0.029\\
GRU                  & 1.7565 $\pm$ 0.210 & 6.4981 $\pm$ 1.102 & 8.2547 $\pm$ 1.122 & 2.025 $\pm$ 0.045 \\
LSTM                 & 1.5480 $\pm$ 0.185 & 4.3283 $\pm$ 0.824 & \textbf{5.8763 $\pm$ 0.842} & 2.040 $\pm$ 0.055  \\
Transformer          & 7.3642 $\pm$ 0.551 & 4.6888 $\pm$ 0.915 & 12.0530 $\pm$ 1.068 & 2.390 $\pm$ 0.135 \\
\hline
FiSTA (seq=1)  & 2.8091 $\pm$ 0.303 & 4.8272 $\pm$ 1.040 & 7.6363 $\pm$ 1.173 & 1.890 $\pm$ 0.030 \\
\textbf{FiSTA (seq=5)}  & 2.8665 $\pm$ 0.325 & 4.2175 $\pm$ 0.562 & 7.0840 $\pm$ 0.772 & \textbf{1.781 $\pm$ 0.014} \\
FiSTA\_NoSpatial & 2.0050 $\pm$ 0.126 & \textbf{3.9653 $\pm$ 0.679} & 5.9703 $\pm$ 0.698 & 1.862 $\pm$ 0.037 \\
FiSTA\_NoTemporal & 1.9993 $\pm$ 0.107 & 5.7616 $\pm$ 0.626 & 7.7609 $\pm$ 0.637 & 1.822 $\pm$ 0.020 \\
FiSTA\_NoFiLM & 2.1463 $\pm$ 0.072 & 9.0212 $\pm$ 3.291 & 11.1674 $\pm$ 3.303 & 1.889 $\pm$ 0.008 \\
FiSTA (seq=8)  & 3.0448 $\pm$ 0.339 & 4.2995 $\pm$ 0.745 & 7.3443 $\pm$ 0.954 & 1.890 $\pm$ 0.038\\
FiSTA (seq=12) & 3.1464 $\pm$ 0.380 & 5.4129 $\pm$ 1.188 & 8.5593 $\pm$ 1.405 & 1.935 $\pm$ 0.010\\
FiSTA (seq=16) & 3.1618 $\pm$ 0.368 & 4.2071 $\pm$ 0.658 & 7.3689 $\pm$ 0.878 & 1.923 $\pm$ 0.018\\
\hline
Jacobian-based numerical IK  & \multicolumn{1}{c}{/} & 5.0818 $\pm$ 0.755 & 5.0818 $\pm$ 0.755 & \multicolumn{1}{c}{/}\\
\hline
\end{tabular}
\label{runtime_com}
\end{table*}

\subsubsection{Runtime Comparison}
To assess the deployment potential of our model in practical applications, we conducted a comprehensive runtime test on an NVIDIA RTX 4070 GPU. The total runtime for each model was calculated by averaging the runtime for one arm across the ACCAD trajectories. In this comparison, the total time is divided into two main stages as shown in Table~\ref{tab:final_runtime_comparison}: (i) Preprocess, which includes the neural network's input/output handling and the model inference itself. (ii) IK, which represents the time taken by the subsequent IK solver. Our optimal model, \texttt{FiSTA (seq=5)}, achieved the best performance, though its overall running time ranked fourth among all the models. The final computation takes 7.0840 ms per step, compared to 5.0818 ms for the Jacobian-based numerical IK, introducing only an additional ~2.0022 ms overhead. Even with this overhead, the runtime still supports a control frequency of approximately 141.2 Hz, which satisfies common real-time control requirements.

\subsubsection{Conclusion} We set the default history length for FiSTA to $L=5$. The network was then trained on the entire dataset for a total of 1 hour and 50 minutes on an NVIDIA RTX 4090 GPU. This final trained model is used for all subsequent elbow-alignment experiments in Section IV-B and IV-C.

\begin{figure}[t]
    \centering
    \includegraphics[width=0.48\textwidth]{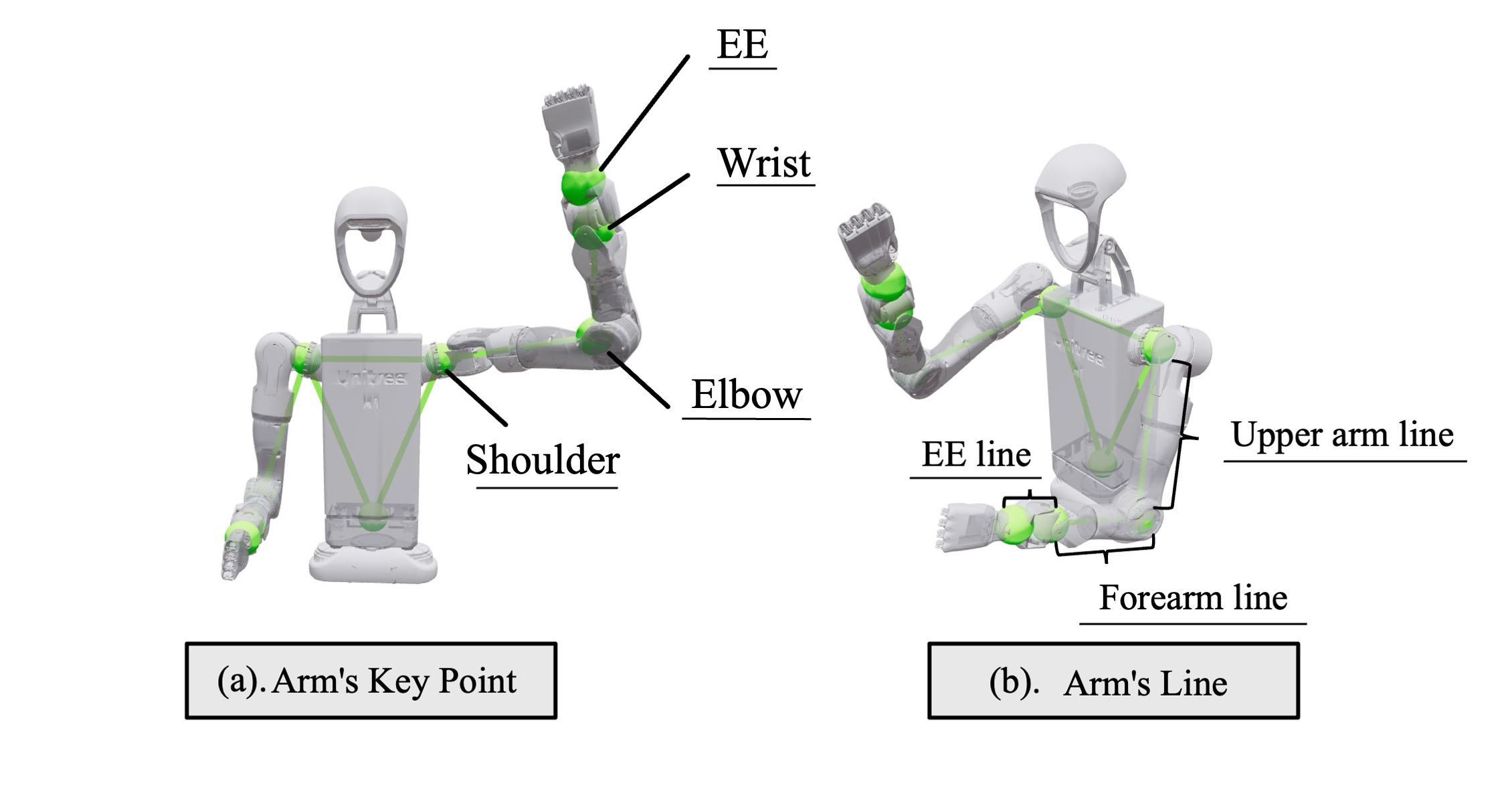}
    \caption{Keypoints and segments for metric computation. Display the four key points (Shoulder, Elbow, Wrist, EE) and three line segments (Upper-arm, Forearm, EE line) used for quantitative indicators of anthropomorphism.}
    \label{fig:eval_intro}
\end{figure}

\begin{figure*}[t]
	\centering
	\includegraphics[width= 0.98\textwidth,height=6.5cm]{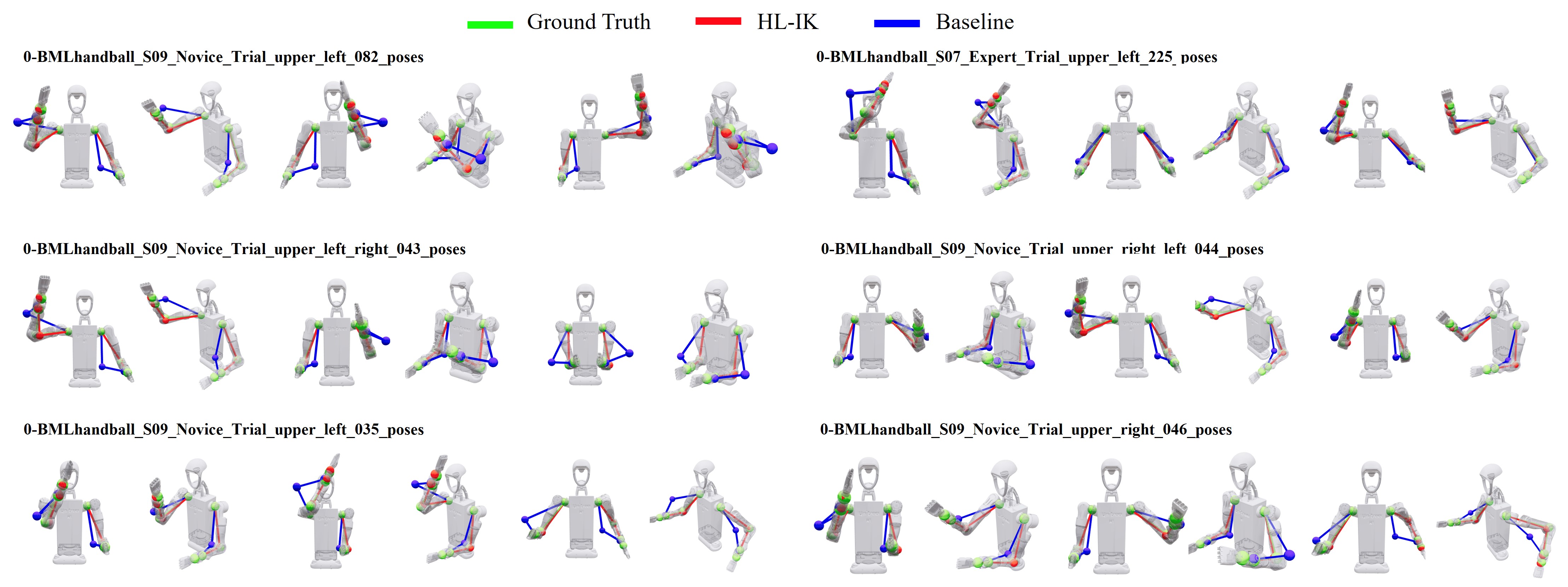}
	\caption{The robot motions. Comparison of robot motions reconstructed by different methods: Ground Truth (green), HL-IK (red), and swivel-angle–based Baseline (blue). In each visualization, four spheres on each arm represent the shoulder, elbow, wrist, and EE, respectively. The gray robot mesh is rendered from the ground-truth poses, while the colored skeletons illustrate the predicted trajectories of different methods.}
\label{sim-traj}
\end{figure*}

\subsection{Comparative Analysis of Elbow Alignment}
When incorporating elbow alignment into IK, we set the weight coefficients as follows.
\begin{subequations}\label{eq:ik-weights}
\begin{align}
&\mathbf{W}_{\mathrm{ee}}     = \operatorname{diag}(50\,\mathbf{I}_3,\,40\,\mathbf{I}_3),\\
&\mathbf{W}_{\mathrm{elbow}}  = \operatorname{diag}(20\,\mathbf{I}_3,\,5\,\mathbf{I}_3),\\
&\mathbf{W}_{\mathrm{smooth}} = 0.35\,\mathbf{I}_d.
\end{align}
\end{subequations}

\noindent{In $\mathbf{W}_{\mathrm{ee}}$ and $\mathbf{W}_{\mathrm{elbow}}$, the first $3{\times}3$ block weights translational residuals and the second weights rotational residuals; $\mathbf{I}_d$ denotes the $d{\times}d$ identity where $d$ is the number of joints.}

For quantitative evaluation, we use a new human-motion dataset (not for network training) - BMLhandball (649 trajectories, 183,806 steps) \cite{AMASS_BMLhandball}. The trajectory generated by the original retargeting method serves as the reference. We compare two IK solvers: (i) a swivel-angle–based baseline, following the method in \cite{Liuc}, and (ii) our elbow-aligned approach, HL-IK. We conduct a comprehensive evaluation using four metrics: two that quantify arm-configuration similarity and two that EE tracking precision.

\subsubsection{Arm's Key Point Position Error} Measures the absolute positional deviation of key arm points $K_\text{arm}= \text{\{elbow, wrist, EE\}}$ from the reference trajectory. This is mainly for calculating the error of key points on the arm configuration after IK solving. Fig.~\ref{fig:eval_intro} (a) demonstrated these arm's key points.
\begin{equation}
E_{\text{arm}}^p(t) = \sum_{k \in K_{\text{arm}}} \left\| \boldsymbol{p}_{k}^{\text{re}}(t) - \boldsymbol{p}_{k}^{\text{ik}}(t) \right\|^2, 
\end{equation}    
where $\boldsymbol{p}_{k}^{\text{re}}(t)$ and  $\boldsymbol{p}_{k}^{\text{ik}}(t)$ represent the 3D positions of the key points from the reference trajectory and the IK solution at time $t$, respectively. 

\subsubsection{Arm' Line Angle Error} Quantifies the angular deviation of $L_{\text{arm}}= \! $ \{upper arm line (shoulder-elbow), forearm line (elbow-wrist), EE line (wrist-EE) \}. This is mainly for calculating the error of the direction of line segments on the arm configuration after IK solving. Fig.~\ref{fig:eval_intro} (b) demonstrated these arm's line.
\begin{equation}
E^\theta_\text{arm}(t) =  \sum_{l \in L_{\text{arm}}} \arccos\left(\frac{\boldsymbol{\upsilon}_{l}^{\mathrm{re}}(t) \cdot \boldsymbol{\upsilon}_{l}^{\mathrm{ik}}(t)}{\left\|\boldsymbol{v}_{l}^{\mathrm{re}}(t)\right\|\left\|\boldsymbol{v}_{l}^{\mathrm{ik}}(t)\right\| + \alpha} \right),
\end{equation}
\noindent
where $\boldsymbol{\upsilon}_{l}^{\mathrm{re}}(t)$ and $\boldsymbol{\upsilon}_{l}^{\mathrm{ik}}(t)$ denote the arm-segment vectors of the reference trajectory and the IK solution at time $t$, respectively. $\alpha>0$ is a small numerical constant added for stability to avoid division by zero.

\subsubsection{EE Position Tracking Error} Evaluates the 3D positional tracking accuracy of the EE relative to the reference trajectory.
\begin{equation}
    E_{\text{ee}}^p(t) = \left\| \boldsymbol{p}_{\text{ee}}^{\text{re}}(t) - \boldsymbol{p}_{\text{ee}}^{\text{ik}}(t) \right\|^2 ,
\end{equation}  
where $\boldsymbol{p}_{\text{ee}}^{\text{re}}(t)$ and $\boldsymbol{p}_{\text{ee}}^{\text{ik}}(t)$ represent EE position of the reference trajectory and the IK solution at time $t$, respectively.

\subsubsection{EE Orientation Tracking Error} Evaluates the 3D orientational tracking accuracy of the EE relative to the reference trajectory.
\begin{subequations}
\begin{gather}
  E_{\text{ee}}^o(t) = {\phi} \left( \mathbf{R}^{\text{re}}_{\text{ee}}(t) \cdot \left(\mathbf{R}^{\text{ik}}_{\text{ee}}(t)\right)^{-1}\right) ^2  , \\
  \phi=\operatorname{acos}\left(\frac{\operatorname{tr}(\mathbf{R})-1}{2}\right)
\end{gather}
\end{subequations}
where $\mathbf{R}^{\text{re}}_{\text{ee}}(t)$ and $\mathbf{R}^{\text{ik}}_{\text{ee}}(t)$ represent the rotation matrices of the EE from the reference trajectory and the IK solution at time t, respectively. $\phi$ quantifies the rotational difference as a specific scalar value representing the angle and $\text{tr}(\cdot)$ calculates the trace of the matrix.

% \begin{table*}[t]
% \centering
% \caption{Comparison of Mean Metrics on the Full Dataset and Challenging Dateset}
% \label{tab:full_dataset_metrics}
% \begin{tabular}{lcccc}
% \hline
% {Quantitative Indicators} & \textbf{HL-IK} & \textbf{Baseline}  & \textbf{HL-IK} (Challenging) & \textbf{Baseline} (Challenging)\\
% \hline\hline
% {Key Point Position Error (m)}   \rule{0pt}{2.5ex}  & $5.944 \times 10^{-2} \pm 0.00412$ & $8.571 \times 10^{-2} \pm 0.00942$   &$5.977 \times 10^{-2} \pm 0.00436$  & $10.338 \times 10^{-2} \pm 0.00812$ \\
% {Line Angle Error (rad)} & $3.565 \times 10^{-1} \pm 0.02438$ & $5.515 \times 10^{-1} \pm 0.08580$  & $3.575 \times 10^{-1} \pm 0.02190$ & $6.791 \times 10^{-1} \pm 0.06978$\\
% {EE Position Error (m)}  & $4.575 \times 10^{-3} \pm 0.00067$ & $1.847 \times 10^{-3} \pm 0.00028$  & $5.405 \times 10^{-3} \pm 0.00218$ & $3.642 \times 10^{-3} \pm 0.00483$\\
% {EE Orientation Error (rad)}     & $8.110 \times 10^{-4} \pm 0.00021$ & $4.621 \times 10^{-4} \pm 0.00008$  & $10.728 \times 10^{-4} \pm 0.00331$ & $9.495 \times 10^{-4} \pm 0.00141$\\
% \bottomrule
% \end{tabular}
% \end{table*}

\begin{table*}[t]
\centering
\caption{Comparison of Mean Metrics on the Full Dataset and Challenging Dataset}
\label{tab:full_dataset_metrics}
\begin{tabular}{lcccc}
\toprule
{Quantitative Indicators} & {HL-IK} & {Baseline}  & {HL-IK (Challenging)} & {Baseline (Challenging)}\\
\midrule
{Key Point Position Error (m)}   & $\mathbf{(5.944 \pm 0.412) \times 10^{-2}}$ & $(8.571 \pm 0.942) \times 10^{-2}$   & $\mathbf{(5.977 \pm 0.436) \times 10^{-2}}$  & $(10.338 \pm 0.812) \times 10^{-2}$ \\
{Line Angle Error (rad)} & $\mathbf{(3.565 \pm 0.244) \times 10^{-1}}$ & $(5.515 \pm 0.858) \times 10^{-1}$  & $\mathbf{(3.575 \pm 0.219) \times 10^{-1}}$ & $(6.791 \pm 0.698) \times 10^{-1}$\\
{EE Position Error (m)}  & $(4.575 \pm 0.670) \times 10^{-3}$ & $\mathbf{(1.847 \pm 0.280) \times 10^{-3}}$  & $(5.405 \pm 2.180) \times 10^{-3}$ & $\mathbf{(3.642 \pm 4.830) \times 10^{-3}}$\\
{EE Orientation Error (rad)}     & $(8.110 \pm 2.100) \times 10^{-4}$ & $\mathbf{(4.621 \pm 0.800) \times 10^{-4}}$  & $(10.728 \pm 33.10) \times 10^{-4}$ & $\mathbf{(9.495 \pm 14.10) \times 10^{-4}}$\\
\bottomrule
\end{tabular}
\end{table*}

% \begin{table}[t]
% \centering
% \caption{Comparison of Mean Metrics on Challenging Trajectories}
% \label{tab:challenging_dataset_metrics}
% \begin{tabular}{lcc}
% \hline
% {Quantitative Indicators} & \textbf{HL-IK} & \textbf{Baseline} \\
% \hline\hline
% {Key Point Position Error (m)}   \rule{0pt}{2.5ex}  & $5.977 \times 10^{-2}$ \pm 0.00436 & $10.338 \times 10^{-2}$ \pm 0.00812 \\
% {Line Angle Error (rad)} & $3.575 \times 10^{-1}$ \pm 0.02190 & $6.791 \times 10^{-1}$ \pm 0.06978 \\
% {EE Position Error (m)}       & $5.405 \times 10^{-3}$ \pm 0.00218 & $3.642 \times 10^{-3}$ \pm 0.00483\\
% {EE Orientation Error (rad)}     & $10.728 \times 10^{-4}$ \pm 0.00331 & $9.495 \times 10^{-4}$ \pm 0.00141 \\
% \bottomrule
% \end{tabular}
% \end{table}

Based on the four metrics mentioned above, we conducted a comparative analysis on the BMLhandball's \cite{AMASS_BMLhandball} human retargeting dataset. First, we calculated the average error across all steps in the trajectory. The first two columns in Table~\ref{tab:full_dataset_metrics} report the mean errors across all 183,806 steps. Each indicator in the table shows the average error for all steps.The arm keypoint position error and line-angle error are reduced by 30.6\% and 35.4\% compared with the baseline. Meanwhile, EE tracking accuracy decreases only slightly, with an increase of 2.728 mm in EE position error and 0.003489 rad in EE orientation error—an acceptable trade-off for teleoperation.

Since our aim is to probe configuration-tracking advantage in our HL-IK, we rank-ordered trajectories by the ``Arm's Key Point Position Error" achieved by the baseline method and selected the top 20\% as ``challenging trajectories" for further evaluation. We then compared the baseline with HL-IK on this subset. As reported in the last two columns  in Table ~\ref{tab:full_dataset_metrics}, and relative to the first two columns, the baseline shows large increases in both arm's key point position error and line angle error, whereas HL-IK maintains nearly the same error levels as on the full set. On this subset, HL-IK reduces arm keypoint and direction errors by 42.2\% and 47.4\%, respectively. Fig.~\ref{sim-traj} visualizes this gap: the baseline (blue) deviates substantially, while HL-IK (red) remains closely aligned with the reference configuration.

\begin{figure*}[t]
	\centering
	\includegraphics[width= \textwidth]{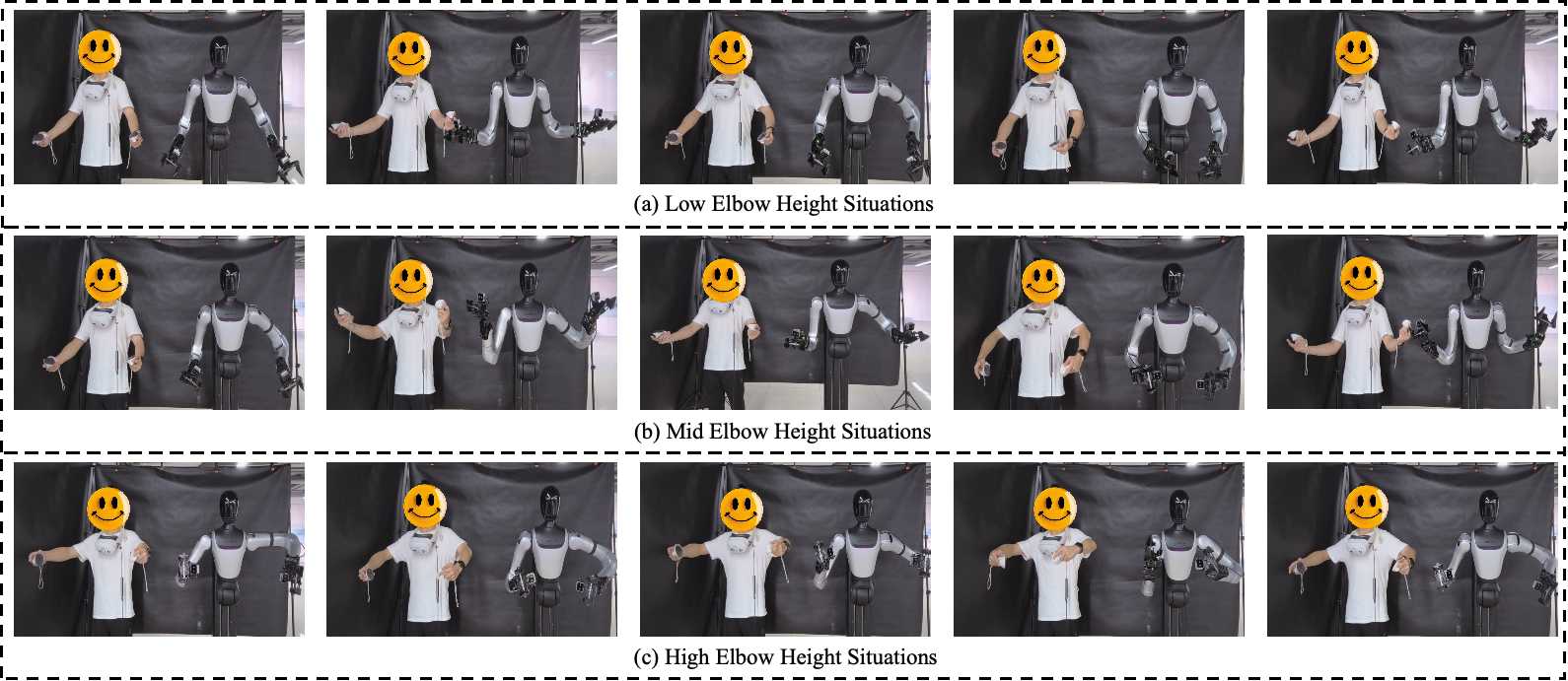}
	\caption{Teleoperation on a physical robot. We deployed the HL-IK framework on the robot’s left arm (shown on the right side of each figure) and a Jacobian-based numerical IK to the right arm, IK solver without elbow-joint optimization on the right arm (shown on the left side of each figure). With HL-IK, the robot not only tracked the end effector but, without any direct sensing, also inferred and aligned with the human elbow position. As a result, its overall arm configuration closely matched the human’s, whereas the standard IK achieved only end effector tracking and failed to reproduce configuration level similarity. Specifically, the higher the elbow joint height, the more apparent the difference. }
\label{phy-traj}
\end{figure*}

\subsection{Teleoperation Experiments on a Physical Robot}
We tested HL-IK on a physical robot that differs substantially from the simulated platform-the upper limbs of a wheeled mobile manipulator. We deployed HL-IK on the left arm and a Jacobian-based numerical IK on the right arm, and qualitatively compared their arm configurations during teleoperation. Our hardware study primarily aims to demonstrate end-to-end teleoperation feasibility and user experience; therefore, we benchmark against a Jacobian-based IK that is widely used in practical teleoperation for robots, rather than the swivel-angle–based IK evaluated in simulation. We use a Meta VR headset to provide the EE target, and no additional sensing is used. As shown in Fig.~\ref{phy-traj}, HL-IK enabled the robot to track the end effector and, without direct sensing, infer and align with the human elbow position. As a result, the robot’s overall arm configuration closely matched the human’s, whereas the baseline IK achieved only end effector tracking and failed to reproduce configuration level similarity. In particular, this difference becomes increasingly apparent with higher elbow joint heights.

\section{Conclusions and Limitations}
In this paper, we introduced HL-IK, a lightweight IK framework that preserves EE tracking while producing human-like arm configurations without full-body sensing. The key idea is a learned elbow prior: a FiSTA network predicts the next-step elbow pose from a short motion history and the desired hand target, and this prediction is injected as a small residual cost alongside EE tracking and smoothness in a standard Levenberg–Marquardt IK stack. This design integrates seamlessly with generic kinematics libraries, adds negligible overhead (best trade-off at a 5-frame history), and significantly improves anthropomorphism. On a 183k-step test, HL-IK reduces arm key-point and directional errors by 30.6\% and 35.4\% on average across all steps—and by 42.2\% and 47.4\% on the most challenging trajectories—while maintaining acceptable EE accuracy. Teleoperation experiments on a physical robot likewise clearly demonstrate the effectiveness of our method in improving arm-configuration similarity in IK.

Despite these encouraging results, our current formulation primarily leverages elbow alignment. While effective in improving overall arm appearance, human-likeness is also influenced by other factors such as shoulder shrugging, torso compensation and habitual wrist posture. Consequently, using only an elbow residual may not consistently match human subjective preferences across all tasks, particularly those requiring complex upper-limb coordination or strong cross-joint coupling. Looking ahead, we will explore richer upper-limb priors, extend to whole-body control, enable online adaptation and personalization, and unify collision and task constraints within a single optimization framework to further improve generalization and robustness in real deployments.

% In this paper, we introduced HL-IK, a lightweight IK framework that preserves  EE tracking while producing human-like arm configurations without full-body sensing. The key idea is a learned elbow prior: a FiSTA network predicts the next-step elbow pose from a short motion history and the desired hand target, and this prediction is injected as a small residual cost alongside EE tracking and smoothness in a standard Levenberg–Marquardt IK stack. This design integrates seamlessly with generic kinematics libraries, adds negligible overhead (best trade-off at a 5-frame history), and significantly improves anthropomorphism. On a 183k-step test, HL-IK reduces arm key-point and directional errors by 30.6\% and 35.4\% on average across all steps—and by 42.2\% and 47.4\% on the most challenging trajectories—while maintaining acceptable EE accuracy. Teleoperation experiments on a physical robot likewise clearly demonstrate the effectiveness of our method in improving arm-configuration similarity in IK. Looking ahead, we will explore richer upper-limb priors, extend to whole-body control, enable online adaptation and personalization, and unify collision and task constraints within a single optimization framework to further improve generalization and robustness in real deployments.

\bibliographystyle{IEEEtran}
\bibliography{ref}

\addtolength{\textheight}{-12cm}   % This command serves to balance the column lengths
                                  % the height of the last page by a suitable amount.
                                  % sure that you do not shorten the height too much.

\end{document}